# Reference Setup for Quantitative Comparison of Segmentation Techniques for Short Glass Fiber CT Data

Tomasz Konopczyński[123], Jitendra Rathore[4], Thorben Kröger[3], Lei Zheng[1], Christoph S. Garbe[2], Simone Carmignato[4], Jürgen Hesser[12]

[1]Experimental Radiation Oncology, Department of Radiation Oncology, University Medical Center Mannheim, Heidelberg University, Germany
[2]Interdisciplinary Center for Scientific Computing (IWR), Heidelberg University, Germany
[3]Volume Graphics GmbH, Heidelberg, Germany
[4]Department of Management and Engineering, University of Padova, Vicenza, Italy

**Abstract**
Comparing different algorithms for segmenting glass fibers in industrial computed tomography (CT) scans is difficult due to the absence of a standard reference dataset. In this work, we introduce a set of annotated scans of short-fiber reinforced polymers (SFRP) as well as synthetically created CT volume data together with the evaluation metrics. We suggest both the metrics and this data set as a reference for studying the performance of different algorithms. The real scans were acquired by a Nikon MCT225 X-ray CT system. The simulated scans were created by the use of an in-house computational model and third-party commercial software. For both types of data, corresponding ground truth annotations have been prepared, including hand annotations for the real scans and STL models for the synthetic scans. Additionally, a Hessian-based Frangi vesselness filter for fiber segmentation has been implemented and open-sourced to serve as a reference for comparisons.

## 1 Introduction

Short-fiber reinforced polymers (SFRP) are used increasingly as a structural material due to the ease of fabrication, economical production, and most importantly their superior mechanical properties. The most common fabrication processes for SFRP composites are injection molding and extrusion compounding. The influence of fiber characteristics (e.g. fiber orientation, fiber length distribution and the percent composition in the final product) on the mechanical properties of SFRP composites is of particular interest and significance for manufacturers [1]. These fiber characteristics are affected by the processing conditions (melt temperature, mold temperature, packing pressure and cooling time); therefore, reliable information about fiber characteristics is much needed for the process optimization during the product development phase. In this work we focus on the (single)-fiber segmentation of volumetric images which is needed for most of the precise local measurements of SFRP composite characteristics. A binary segmentation of fibers/non-fibers leads to a direct measurement of the fiber ratio in a volume. Moreover, a single-fiber segmented volume can be used to retrieve the correct length and orientation distribution in a volume.

A number of methods used for fiber segmentation are described in literature [7, 8, 10]. However, it is difficult to compare their performance. In order to judge the suitability of a given algorithm for the task of fiber segmentation, it is necessary to have reliable measures for the quality and accuracy of different methods. To address this problem we propose a reference dataset on which a quantitative comparison of segmentation techniques can be performed. Following the "Grand Challenge" [3] concept used in the medical imaging community, we host the data in a similar framework. We provide a publicly available standard dataset of SFRP composite CT scans together with its ground truth and a set of metrics, with which different fiber segmentation techniques may be verified. We used x-ray scans of micro-injection molded parts from a commercially used SFRP material for a case study. To support the hand-annotations, we developed a post-processing method for ground truth preparation. Even with the use of our post-processing method, hand-annotating the data is very time-consuming and results in only a low number of annotated fibers. Therefore, we have additionally created a computational model of SFRP composites. By using the model, we have performed a number of synthetic CT scans, which are provided in the dataset. The synthetic scans are similar to the real data, and thus the same set of algorithms might be verified on them. For the evaluation criteria we use the Dice coefficient [4] and Adjusted Rand Index [5], which are standard metrics commonly used to validate the quality of segmentations techniques. The dataset containing the experimental scans, the synthetic scans and corresponding ground truths together with the algorithms is available at http://ipm-datasets.iwr.uni-heidelberg.de.

## 2 Related work

There are many reference datasets available to compare different methods for various task specific applications. The framework of the "Grand Challenge" organized by the Consortium of Open Medical Image Computing is very popular in the field of image processing. At the moment of writing this publication it is currently hosting 134 individual challenges [3] (i.e. annotated datasets together with evaluation metrics and results of different algorithms). One example of a successful and similar challenge to ours is the "VESsel SEgmentation in the Lung 2012 challenge" [2] in which the participants were asked to segment vessels from CT thorax scans. Due to the similar tubular-looking structure of vessels, the algorithms for fiber segmentation are very similar to those for vessel segmentation. Therefore, some of the described methods below were primarily designed for vessel segmentation.

One of the easiest and most widely used algorithms for segmentation is Otsu [6], which is a gray value based histogram method. The reliability of histogram-based methods is limited due to noise and brightness variations over the image. To overcome this problem, slice-wise circular Hough Transform or Circular Voting [7, 8] can be used. These methods take the advantage of the geometrical information about the scanned part, e.g. the radius of a tubular structure, in order to search for circular or elliptical structures in 2D slices. However, these algorithms do not scale well to 3D data. The most common methods to segment tubular looking structures in 3D data use Hessian based filters. One of the first Hessian based methods, Frangi vesselness filter [9], was initially developed for segmenting vessels in biomedical images and is now commonly used as a preprocessing step on 3D CT data. For fiber-reinforced polymers, a priori information such as the radius of fibers or expected orientation distribution can be incorporated into the algorithm. The method developed by Zauner et al. [10] is a good example of using a priori knowledge dedicated to the analysis of fiber-reinforced polymers. There is no standard technique for quantitative determination of how well the methods actually work on the SFRP CT scans.

Creating a model of a specimen is a common method of evaluating an algorithm. A similar approach has been proposed to evaluate 3D fiber orientation algorithms based on x-ray computed tomography models of SFRP [11]. This work however, omits the x-ray imaging simulations. Similarly, Z. Bliznakova et al. proposed a computational model for carbon fiber-reinforced polymers [12]. The model was later used in an entire pipeline for carbon fiber-reinforced polymers modeling for x-ray imaging simulation together with a CT scan simulation [13]. Unfortunately, to the best knowledge of the authors none of the models is publicly available. In our work, we first create a geometrical model of SFRP composite, then simulate an x-ray imaging and perform CT reconstruction of the simulated projections. Our simulation is also set up to match the parameters of the material used for the experimental scans of the SFRP specimen on a Nikon MCT225 X-ray CT system.

## 2 Dataset preparation

We provide a new public dataset for evaluating (single)-fiber segmentation techniques. The provided scans exhibit typical artifacts and limited resolution. We decided to divide the dataset into two main parts with one part containing the real experimental CT scans of SFRP composites and the other containing synthetic CT scans. The scans are acquired (or modeled) at both high resolution (HR) and low resolution (LR). For now, only the real HR CT scans have been hand-labeled. The annotations are further post-processed in order to create the segmentation ground truth. The synthetic data contains simulated scans of SFRP by using computational model. Therefore, for the synthetic volumes a perfect ground truth is known. Both types of data are provided with training and test sets with corresponding labels (post-processed hand annotations for the real data and STL models for the simulated data). The annotations are only available for the training datasets, whereas the annotations for the test data are kept hidden from the public. The algorithms may be trained and initially verified on the training dataset, but the final score is computed and evaluated on the organizers' side with the use of the hidden ground truth and labels. Example slices of the data can be seen in Figure 1 (a) and (b).

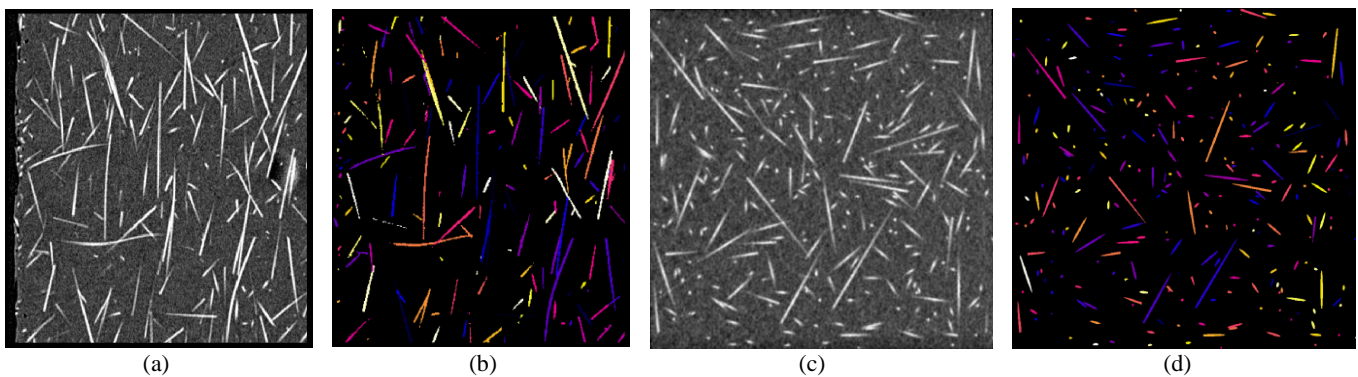

(a) (b) (c) (d)

Figure 1: Example slices of the data together with its corresponding ground truth (the random coloring has been added for visualisation purposes. a) Slice from a real HR experimental data, b) corresponding hand made annotated segments after post-processing, c) slice from a synthetic HR data, d) corresponding slice of the STL model.

### 2.1 Real experimental data

The part used for our dataset were manufactured by micro injection molding using PBT-10% GF, a commercial polybutylene terephthalate PBT (BASF, Ultradur B4300 G2) reinforced with short glass fibers (10% in weight). The fibers in the material have a diameter of 10-14 µm and are approx. 1.1 mm in length. We acquired the experimental scans on a Nikon MCT225 X-ray CT system. The scans are performed in both HR and LR with isotropic resolution and the corresponding voxel sizes are 3.9 µm and 8.3 µm respectively. The HR data has been further hand annotated by use of the Knossos labeling tool [14]. During the annotation process, each fiber gets a different index, which is a unique ID number. The annotation process took around an hour per 200 fibers per annotating person. This includes annotation of particles, which may be used as false positives examples in the future. Each fiber or a particle is annotated as a set of connected points (i.e. one polygonal chain per fiber). Currently we provide 8,000 fiber annotations for two regions of two scans in HR (around 4,000 annotations per volume). The website is being updated with more data and annotations.

As the annotations from Knossos are single connected points, we developed a post-processing script to convert it into a volume of annotated segments. The steps of the pipeline are as follows. First, the 3D version of Bresenham's line algorithm [15] is used on the hand-labeled polygonal chains. The algorithm draws straight lines on a 3D grid and renders them into a volume. In the next step the lines are used as seeds for the 3D region growing segmentation algorithm [16] which is applied on the input volume (real volumetric data). The region growing stopping criteria has to be set manually and the value depends on the data. The resulting output is the desired volume of annotated segments. In this way, each experimental scan has a corresponding volume with ground truth voxel labels. That is, the annotated voxels have corresponding index value for voxels assumed to contain fibers or value 0 for the background (not fibers). Figure 2 presents example slices from the pipeline correspondingly. We have implemented Bresenham's line algorithm in Python and used the VIGRA implementation of the 3D region growing algorithm [17].

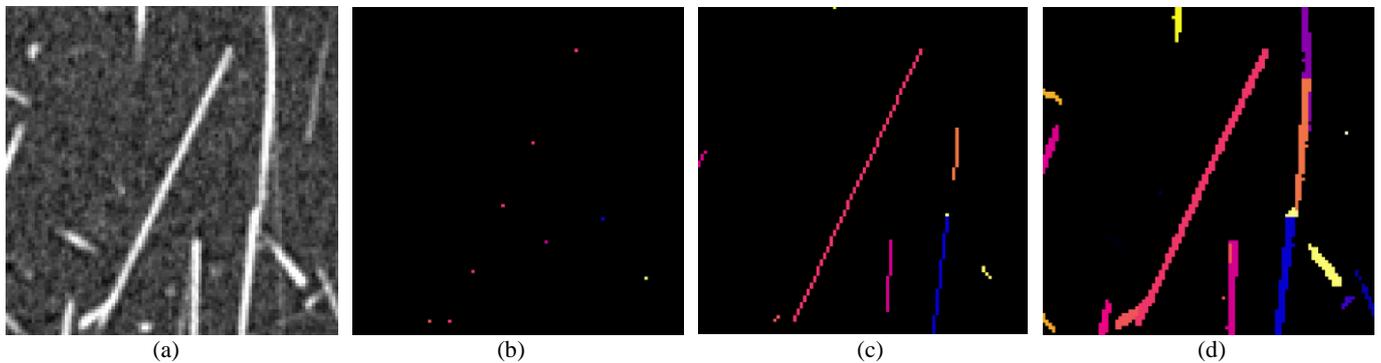

Figure 2: The post-processing pipeline steps. For the visualisation purpose each fiber has a random color. a) Zoomed region from a slice of real volumetric data, b) the Knossos polygonal chains rendered as voxels, c) rendered lines after 3D Bresenham's line algorithm, d) rendered annotated segments after 3D region growing.

## 2.2 Synthetic data

Since the hand-annotated data is limited, we can greatly extend the number of volumes in the data; and the quality of the hand-annotations may raise a discussion. Therefore, adding synthetic scans to this study provides a number of advantages. First, there are more annotated volumes without the need of manual annotations. Second, having the computational model allows us to have the perfect ground truth for the synthetic scans in comparison to hand-annotations on the real scans. The synthetic scans are based on a computational SFRP composite model generator, which has been written by use of an in-house software. We wanted our synthetic scans to have similar parameters to the real scans. The settings of the model were set to mimic the characteristic of the PBT-10%GF material [18]. For the x-ray imaging simulation of a scan, we have used the software package aRTist 2 [19]. We designed the source and the detector to match the parameters of the Nikon MCT225 X-ray CT system [20].

The proposed computational model of SFRP composite has been set to match the fiber content of 5.4±0.5% of the entire volume (that is around 10% weight fraction). We create the SFRP composite model by defining each fiber in the model as a thin cylinder with a fixed radius at a random orientation, position and with a random length. For this work, we set the volume of the composite to be a $(2\text{ mm})^3$ cube. The radius has been set to 6.5 µm and the length is sampled from a normal distribution with mean of 500±100 µm. The algorithm for the model creation works in an iterative way. Once the desired dimension of the synthetic volume is defined (the dimensions of a cube), the algorithm tries to fit randomly created fibers inside it. At the end of each successful iteration (i.e. an iteration after which a fiber is added to the model), the fiber ratio is calculated based on the provided densities of the fiber and epoxy material [21]. For instance, in the example in Table 1, out of 150,000 randomly generated fibers only 6,628 fit into the cube. The algorithm stops when the desired fiber ratio or the maximum number of attempts is reached. If the fiber has been generated inside the cube, and fits in such a way that it does not overlap with the previous fibers, it is saved as a set of points defining its surface. Because of this fitting process, the fibers are forced to be almost parallel to the surface of the cube the closer they are to its surface. Computation time depends on the properties of the model. The higher the desired content ratio of fibers the longer it takes to create the model. For the properties described above, preparation of one computational model takes around 6 hours on a single CPU. Example parameters for a single model and resulting statistics are presented in Table 1. The length and orientation distributions of the resulting example model created by use of the parameters from Table 1 are shown in Figure 3.

For the x-ray imaging simulations, we use the generated STL model of fibers and embeds it inside a cube geometry, which is used as epoxy matrix. The densities of glass fibers (2.54 g/cc) and the surrounding epoxy matrix (1.31 g/cc) are set to the characteristics of the PBT-10%GF material [22]. The synthetic scans are also simulated with isotropic resolution in LR (8.3 µm) and HR (3.9 µm). The source-object distance (SOD) and source-detector distance (SDD) were set to 23.36 mm and 1177.08 mm for the HR and 48.96 mm and 1176.96 mm for the LR respectively. In both HR and LR settings, we have used four projections to average and performed in total 2,000 projections per model. The detector and source parameters are the same for LR and HR settings and are similar to the characteristics of the Nikon MCT225 X-ray CT system. The detector size has been set to 2048 × 2048 pixels, with pixel resolution of 0.2 mm. The voltage has been set to 120 kV, and the current to 71 µA. For the noise factor, we have checked the signal to noise ratio (SNR) of the air of a real experimental projection and matched it to our simulation. The reconstruction is performed by use of the standard CT filtered back projection (FBP) reconstruction algorithm implemented in the CT Reconstruction module of the commercial software VGStudio Max 3.0 [23]. Fig. 1 (c) and (d) show one slice of the synthetic scan and the corresponding STL model. The STL model serves as a ground truth for the resulting volume. In order to get a binary mask the STL models are rendered into binary volumes. The entire simplified process of synthetic volume preparation from the computational STL model is presented in Figure 4.

| Computational model parameters | |
|---|---|
| Box dimension | 2×2×2 mm$^3$ |
| Mean length | 500 µm |
| Length deviation | 100 µm |
| Radius | 6.5 µm |
| Number of attempts | 150000 |
| **Resulting model statistics** | |
| Number of fibers | 6628 |
| Max length of a fiber | 898.48 µm |
| Min length of a fiber | 126.16 µm |
| Entire volume | 8,000,000,000 µm$^3$ |
| Fiber content volume | 431,941,002 µm$^3$ |
| Fiber volume fraction | 5.40% |

Table 1: Synthetic model parameters and resulting STL model statistics.

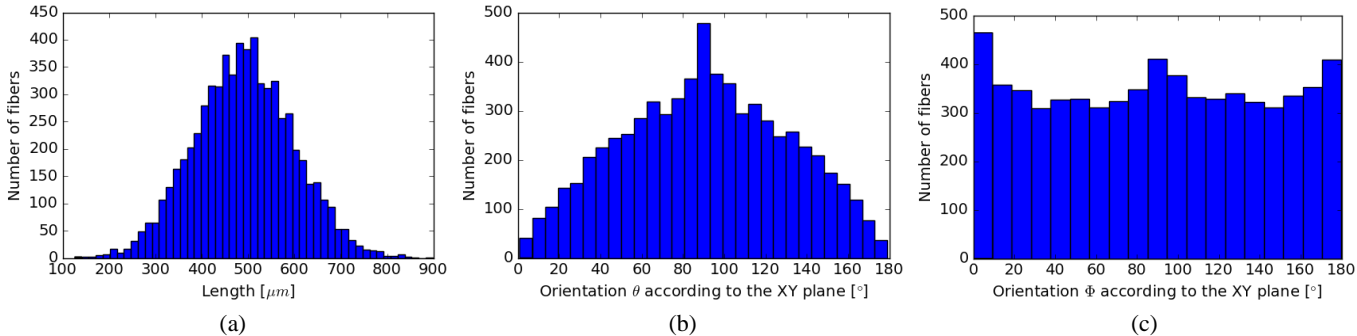

(a) (b) (c)

Figure 3: Resulting distributions of an example synthetic computational STL model with 5.4% fiber volume ratio. a) Fiber length distribution, b) theta orientation distribution and c) phi orientation distribution according to the XY plane.

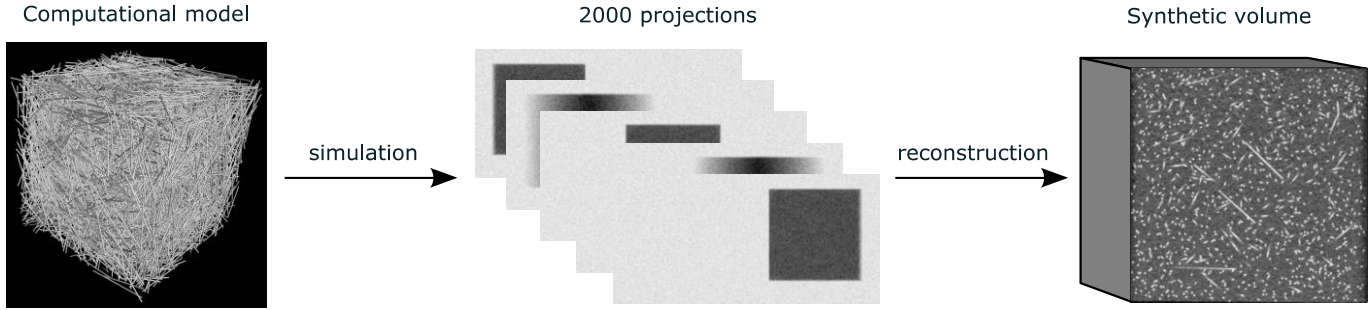

Figure 4: Sketch of the synthetic volume preparation. The computational model of SFRP in STL format is used in combination with the commercial simulated x-ray imaging software to produce a number of projections. The projections are reconstructed with the standard FBP algorithm implemented in VGStudio Max.

## 4 Reference algorithm

As a baseline reference, we propose to use a tubular structure enhancement filter [24]. Concretely a Frangi vesselness filter [9], which derives structural information from the Hessian eigenvalues $|\lambda_1| \leq |\lambda_2| \leq |\lambda_3|$. For computation of the Hessian, a number of Gaussian second derivatives at scales $\sigma_s$ are used. The use of Gaussian kernels allows using separable convolution, which speeds up the computation time. The Gaussian second order derivative of image $I$ at scale $\sigma$ and point $x$ is given by

$$\frac{\partial^2 I_\sigma}{\partial x^2} = I(x) \frac{\partial^2 G(\sigma, x)}{\partial x^2}$$

In [9], the authors proposed three measures to distinguish structures in the images. The Frangi vesselness filter $V(x)$ uses a combination of these measures, and is given by

$$V(\sigma, x) = \begin{cases} 0 & \text{if } \lambda_2, \lambda_3 > 0, \\ \left(1 - \exp\left(-\frac{R_A^2}{2\alpha^2}\right)\right) \cdot \exp\left(-\frac{R_B^2}{2\beta^2}\right) \cdot \left(1 - \exp\left(-\frac{S^2}{2c^2}\right)\right) & \text{otherwise} \end{cases}$$

where the first measure $R_B = \frac{|\lambda_1|}{\sqrt{|\lambda_2 \lambda_3|}}$ quantifies deviation from a blob-like structure, the second $R_A = \frac{|\lambda_2|}{|\lambda_3|}$ quantifies the difference between plate-like and line-like structures and the third $S = \|H\|_F = \sqrt{\sum_i \lambda_i^2}$ quantifies the presence of the background noise. $\alpha, \beta, c$ are real valued positive parameters of the filter. The filter is finally embedded in a multi-scale framework by the following formula

$$V(x) = \max_\sigma V(\sigma, x)$$

As a baseline reference, we use the 3D version of the Frangi vesselness filter implementation in Python. Additionally, for the local orientation distribution, a structure tensor based algorithm based on the 2D implementation of ImageJ plugin OrientationJ [25] has been implemented and extended to 3D. It may serve as a help tool to evaluate additional statistics of the datasets.

## 5 Evaluation criteria

For evaluation criteria, we propose to use the two following common metrics [4, 5]. One for the binary "fiber/non-fiber" segmentation task and one for the single-fiber segmentation task. For the binary classification, we propose to use the mean Dice Coefficient [4]. One can use this metric to compare the predicted segmentation on a pixel level with the ground truth (labeled volumes). Defining the binary ground truth labels as a cluster $B$ and the corresponding predicted binary labels as a cluster $B'$, the Dice coefficient index $D$ is defined by

$$D(B, B') = \frac{2 \times |B \cap B'|}{|B| + |B'|} = \frac{2 \times TP}{2 \times TP + FN + FP}$$

where the intersection operation is the voxel-wise minimum operation, and $|\cdot|$ is the integration of the voxel values over the complete image, $TP$ is the true positive, $FP$ false positive and $FN$ false negative.

For the single-fiber segmentation, we use the Adjusted Rand Index [5]. Defining the ground truth labels as a cluster $C = \{C_1, \ldots, C_k\}$ and the corresponding predicted labels as a cluster $C' = \{C'_1, \ldots, C'_l\}$, the Adjusted Rand Index $R_a$ is of a form:

$$R_a(C, C') = \frac{\sum_{i=1}^k \sum_{j=1}^l \binom{m_{ij}}{2} - t_3}{\frac{1}{2}(t_1 + t_2) - t_3}$$

where $m_{ij} = |C_i \cap C'_j|$, $t_1 = \sum_{i=1}^k \binom{|C_i|}{2}$, $t_2 = \sum_{j=1}^l \binom{|C'_j|}{2}$, $t_3 = \frac{2 t_1 t_2}{n(n-1)}$, and $n$ is the number of voxels in the volume.

Both the Dice Coefficient and the Rand Index vary from 0 to 1, where 1 means a perfect match between the algorithm output and the ground truth mask.

For the baseline method, we have implemented and open sourced the 3D Frangi veselness filter [9]. For instance, our 3D Frangi vesselnes filter implementation achieves a score of 0.704 for the task of binary segmentation on one of the synthetic volumes.

## 4  Conclusions

We provide a dataset of SFRP composite CT scans for quantitative comparison of segmentation techniques together with evaluation metrics. The dataset was designed and prepared in order to evaluate (single)-fiber segmentation algorithms. However, both its real and synthetic part may be used outside the framework of the challenge. For instance, one can use the synthetic data and its corresponding model to evaluate algorithms, which measure the local orientation distribution. For the real part, we have designed and implemented a post-processing pipeline for the Knossos – polygonal chain annotations. For the synthetic part, we designed a computational model of SFRP and used it in combination with third-party computer simulated x-ray imaging software. Currently the ground truth for the real experimental data are available for the HR data only. In the future we plan to extend it to the LR data and update it on the website. Additionally, we proposed a base-line algorithm for the task of fiber segmentation and its implementation in Python. We plan to extend and evaluate our proposed methods for this work in more detail. The post-processing pipeline might be extended and used on its own to measure statistics like length or orientation distribution for a small subset area of a CT scan. In the future, depending on the interest in our dataset, we plan to make a quantitative summary.

**Acknowledgements**

This work was supported by funding from the European Union's Seventh Framework Programme (FP7/2007-2013) under grant agreement PITN-GA-2013-607817-INTERAQCT (INTERAQCT: International Network for the Training of Early stage Researchers on Advanced Quality control by Computed Tomography).

**References**

[1]  S.Y. Fu, B. Lauke, Effects of fiber length and fiber orientation distributions on the tensile strength of short-fiber-reinforced polymers. Composites Science and Technology 56, no. 10: 1179-1190, 1996.
[2]  R.D. Rudyanto, et al., Comparing algorithms for automated vessel segmentation in computed tomography scans of the lung: the VESSEL12 study, Medical image analysis, 18(7), pp.1217-1232, 2014.
[3]  Consortium for Open Medical Image Computing. http://grand-challenge.org/. Accessed: 2016-12-15.
[4]  K. Hameeteman, et al. Evaluation framework for carotid bifurcation lumen segmentation and stenosis grading, Medical Image Analysis 15.4: 477-488, 2011.
[5]  H. Lawrence, P. Arabie, Comparing partitions, Journal of classification 2.1: 193-218, 1985
[6]  N. Otsu, A threshold selection method from gray-level histograms. Automatica 11.285-296: 23-27, 1975.
[7]  T. Fast, A.E. Scot, B.N. Cos, Topological and Euclidean metrics reveal spatially nonuniform structure in the entanglement of stochastic fiber bundles. Journal of Materials Science 50.6: 2370-2398, 2015.
[8]  P. Pinter, B. Bertram, K.A. Weidenmann, A Novel Method for the Determination of Fibre Length Distributions from μCT-data, Conference on Industrial Computed Tomography (iCT), 2016.
[9]  A.F. Frangi, W.J. Niessen, K.L. Vincken, M.A. Viergever, Multiscale vessel enhancement filtering. International Conference on Medical Image Computing and Computer-Assisted Intervention. Springer Berlin Heidelberg, 1998.
[10]  H. Zauner, et al. 3D image processing for single fibre characterization by means of XCT. Acta Stereologica, 2015.
[11]  T. Riedel. Evaluation of 3D fiber orientation analysis based on x-ray computed tomography data. Conference on Industrial Computed Tomography (iCT), 2012.
[12]  K. Bliznakova, et al., Modelling of small CFRP aerostructure parts for X-ray imaging simulation. International Journal of Structural Integrity, 5(3), pp.227-240, 2014.
[13]  K. Tigkos, et al., Simulation study for optimization of X-ray inspection setup applied to CFRP aerostructures, Conference on Industrial Computed Tomography (iCT), 2014.
[14]  M. Helmstaedter, K.L. Briggman, W. Denk, High-accuracy neurite reconstruction for high-throughput neuroanatomy. Nature neuroscience, 14(8), 1081-1088, 2011. https://knossostool.org/. Accessed: 2016-12-15.
[15]  J.E. Bresenham, Algorithm for computer control of a digital plotter. IBM Systems journal, 4(1), pp.25-30, 1965.
[16]  L. Vincent, P. Soille, Watersheds in digital spaces: an efficient algorithm based on immersion simulations. IEEE transactions on pattern analysis and machine intelligence, 13(6), pp.583-598, 1991.
[17]  U. Koethe, Computer Vision Library Vigra. https://github.com/ukoethe/vigra. Accessed: 2016-12-15.
[18]  S.Y. Fu, B. Lauke, Y.W. Mai, Science and engineering of short fibre reinforced polymer composites. Elsevier, 2009.
[19]  C. Bellon, et al. Radiographic simulator aRTist: version 2. Proc. of 18th World Conference on Nondestructive Testing. Durban, South Africa, 2012.
[20]  Nikon MCT225, http://www.nikonmetrology.com/en_EU/Products/X-ray-and-CT-Inspection/Metrology-CT/MCT225-for-Metrology-CT-Absolute-accuracy-for-inside-geometry/(brochure). Accessed: 2016-12-15.
[21]  J.F. O'Gara, G.E. Novak, M.G. Wyzgoski, Predicting the tensile strength of short glass fiber reinforced injection molded plastics. Proceedings of the 10th-Annual SPE® Automotive Composites Conference & Exhibition (ACCE), Troy, MI, USA (pp. 15-16), 2010.
[22]  M. Kinsella, D. Murray, D. Crane, J. Mancinelli, M. Kranjc, Mechanical properties of polymeric composites reinforced with high strength glass fibers, International SAMPE Technical Conference (Vol. 33, pp. 1644-1657), 2001.
[23]  VGStudio Max 3.0, http://www.volumegraphics.com/en/products/vgstudio-max/basic-functionality. Accessed: 2016-12-15.

[24] C. Xiao, et al., A strain energy filter for 3D vessel enhancement with application to pulmonary CT images, Medical image analysis 15.1: 112-124, 2011.
[25] Z. Püspöki, M. Storath, D. Sage, M. Unser, Transforms and Operators for Directional Bioimage Analysis: A Survey. In Focus on Bio-Image Informatics (pp. 69-93). Springer International Publishing, 2016.